\documentclass[final,5p,times,twocolumn, authoryear]{elsarticle}

\usepackage{amssymb}
\usepackage{lipsum}
\usepackage[singlelinecheck=false]{caption}
\usepackage{makecell}

\usepackage{hyperref}
\hypersetup{colorlinks,allcolors=black}

\makeatletter
\def\ps@pprintTitle{%
 \let\@oddhead\@empty
 \let\@evenhead\@empty
 \def\@oddfoot{}%
 \let\@evenfoot\@oddfoot}
\makeatother

\begin{document}

\begin{frontmatter}

\title{The effect of source disclosure on evaluation of AI-generated messages: A two-part study}

\author[a]{Sue Lim\fnref{label1}}
\affiliation[a]{organization={Department of Communication, Michigan State University},%Department and Organization
            addressline={404 Wilson Rd.}, 
            city={East Lansing},
            postcode={48824}, 
            state={MI},
            country={USA}}
\author[a]{Ralf Schmälzle}
\fntext[label1]{Corresponding Author. Email: limsue@msu.edu}

\begin{abstract}

Advancements in artificial intelligence (AI) over the last decade demonstrate that machines can exhibit communicative behavior and influence how humans think, feel, and behave. In fact, the recent development of ChatGPT has shown that large language models (LLMs) can be leveraged to generate high-quality communication content at scale and across domains, suggesting that they will be increasingly used in practice. However, many questions remain about how knowing the source of the messages influences recipients’ evaluation of and preference for AI-generated messages compared to human-generated messages. This paper investigated this topic in the context of vaping prevention messaging. In Study 1, which was pre-registered, we examined the influence of source disclosure on people’s evaluation of AI-generated health prevention messages compared to human-generated messages. We found that source disclosure (i.e., labeling the source of a message as AI vs. human) significantly impacted the evaluation of the messages but did not significantly alter message rankings. In a follow-up study (Study 2), we examined how the influence of source disclosure may vary by the participants’ negative attitudes towards AI. We found a significant moderating effect of negative attitudes towards AI on message evaluation, but not for message selection. However, for those with moderate levels of negative attitudes towards AI, source disclosure decreased the preference for AI-generated messages. Overall, the results of this series of studies showed a slight bias against AI-generated messages once the source was disclosed, adding to the emerging area of study that lies at the intersection of AI and communication.
\end{abstract}


\end{frontmatter}

\section{Introduction}

\textit{“Imagine a world where persuasive content is crafted so masterfully that it becomes nearly indistinguishable from human creation, yet is generated by machines at the click of a button. This groundbreaking study unveils the potential of leveraging large language models (LLMs) to generate compelling messages, and puts it to the ultimate test: can they outperform human-crafted tweets in captivating the minds of their audience?"} (Generated by GPT4 powered ChatGPT).

Recent technological breakthroughs in neural network modeling have ushered in an era of artificial intelligence (AI), and new AI-based systems, such as OpenAI’s ChatGPT, are gaining rapid adoption. Within this context, the term AI generally refers to a field of study that aims to understand and build intelligent machines \citep{luger2005artificial, mitchell2019artificial, russell2021artificial}. The precise and specific definition of intelligence differs based on the approach taken by the researchers, but a common theme is that machines can exhibit cognitive capacities such as intelligence, language, knowledge, and reasoning, which had traditionally been limited to human brains. AI technologies like ChatGPT, or similar systems (e.g., Google’s Bard, Meta’s Llama) are driven by large language models (LLMs), a specific kind of transformer-based neural networks trained on massive amounts of text. Importantly, these LLMs can not only process and categorize text, but they can also be used to generate text that mimics the flow of natural human language \citep{bubeck2023sparks, hirschberg2015advances, wei2022chain}. 

As the above content from ChatGPT shows, LLMs have advanced to the point where even with minimum instructions, they can generate high-quality creative and informative content. This has opened ample opportunities for health researchers and practitioners to leverage LLMs to augment their work. For instance, within health communication, researchers have found that messages generated by LLMs were clear and informative, and exhibited argument strength \citep{karinshak2023working, lim2023artificial, schmalzle2022harnessing, zhou2023ai}. As LLMs continue to expand in these capabilities \citep{bubeck2023sparks}, we can expect to see LLMs being used as tools for generating persuasive health messages. However, the rise of AI-generated content in the public communication environment raises the pressing question of how people react to AI as message creators.

Though this is a relatively novel area of study, there are two relevant bodies of literature that we can draw from: interdisciplinary research about the general sentiment of hesitancy towards novel technologies and source effects research within communication research. It is well-documented that new technologies are often met with skepticism. Studies suggest a general sentiment of hesitancy \citep{von2021transparency} and mild to moderate aversion \citep{castelo2021conservatism, jussupow2020why} towards AI and computer algorithms more broadly. Also, when told that AI was involved in the creation of communicative content, there was some reporting of preference against or lower evaluation of that content (e.g., Airbnb profile writing; \cite{jakesch2019ai-mediated}; email writing; \cite{liu2022will}; generated paintings; \cite{ragot2020ai}; music creation; \cite{shank2023ai}; translation of written content; \cite{asscher2023human}). Within health contexts especially, some studies show that people tend to prefer human practitioners over AI-based technologies like chatbots when receiving consultation about health conditions \citep{miles2021health}, citing lack of personalization and incompetence in addressing individual needs as some of the reasons for hesitancy \citep{longoni2019resistance}.

Second, source effects have been studied extensively in persuasion and communication. For instance, a plethora of literature has examined the influence of various aspects of the source, such as credibility, trustworthiness, and similarity, on people’s attitudes and behavior \citep{okeefe2015persuasion, pornpitakpan2004persuasiveness, wilson1993source}. With the advancement of technology, research also examined source effects in online settings \citep{ismagilova2020effect, ma2017user}. In addition, some of the most well-known theories within communication have examined cognitive mechanisms of source effects (ELM; \cite{petty1986elaboration}; HSM; \cite{chen1999heuristic}). Speaking broadly, the results from these studies show that people’s thoughts about the source of the message shape how they evaluate the communication content from the source. Since there’s already been evidence that LLMs have the potential to be powerful tools in expanding health communication theory and augmenting health campaign practice, it is thus important to investigate how people’s perception of AI influences people’s evaluation of health campaign messages. Moreover, it will also be critical to identify potential moderators of such influence. 

This paper presents two experimental studies that shed light on the influence of source disclosure on the evaluation of prevention messages (see Figure 1). For the first study (study 1), we conducted an experimental study examining how source disclosure influenced people’s evaluation of (in terms of effects perception) and preference for (in terms of ranking) prevention messages generated by a LLM compared to humans. Then a follow-up study (study 2) inspected how the influence of source disclosure varied on the basis of people’s general attitudes toward AI. The findings from our studies have the potential to augment source effects theory within mediated health communication by highlighting how people’s awareness of LLM’s role in message generation influences their evaluation of the messages.

\begin{figure}[hbt!]
	\includegraphics[width=0.5\textwidth]{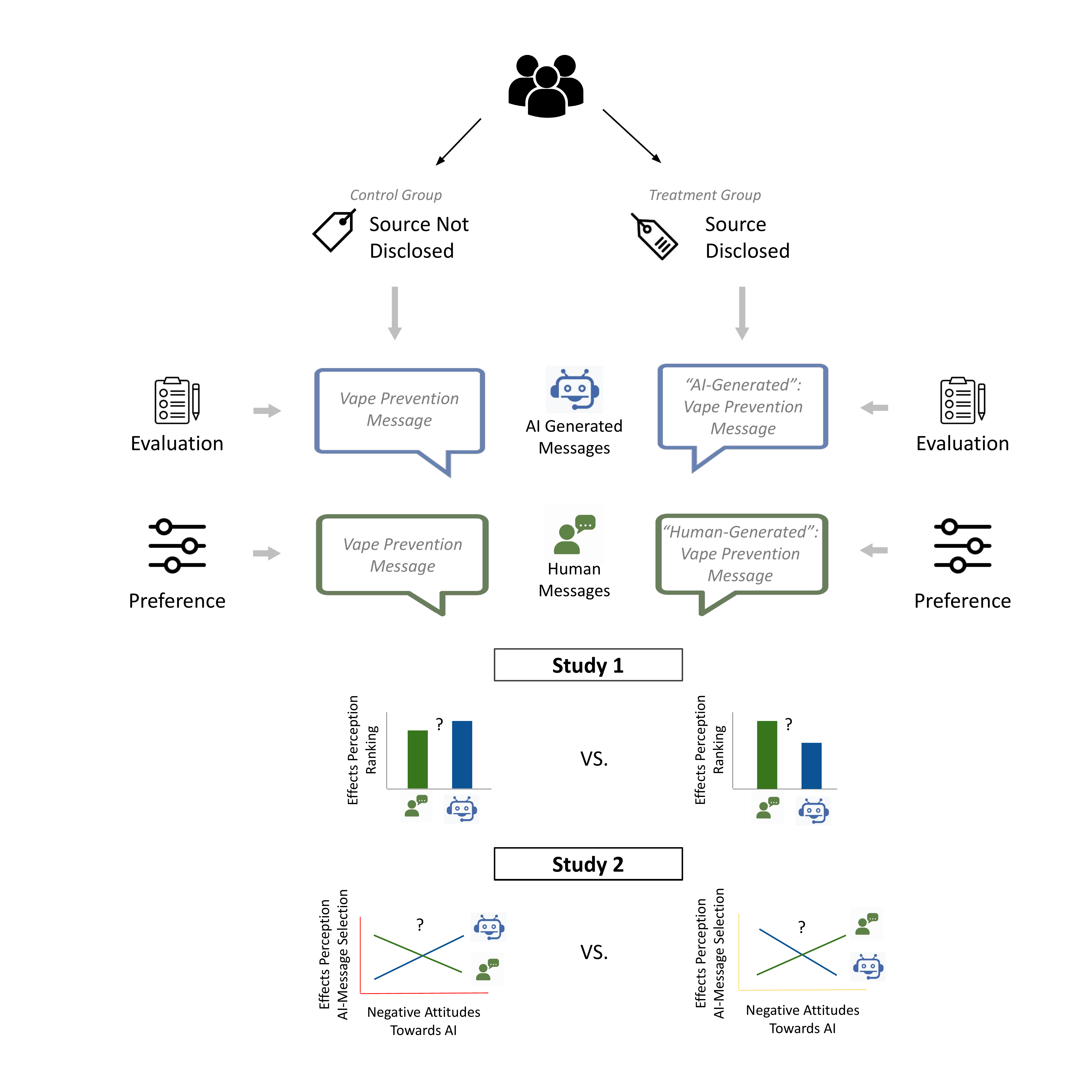}	
	\caption{Conceptual Diagram of Study Design} 
	\label{fig_mom0}%
\end{figure}

\section{Study 1}

The goal of our first study was to examine whether source disclosure influenced people’s evaluations of AI-generated messages as well as their preference for AI as the source of health information. We selected vaping prevention as a health context to examine the evaluation of messages coming from AI source \footnote[1]{Going forward, one could also determine whether the specific health topic matters. For instance, based on psychometric models of risk perception \citep{slovic1987psychometric}, one could predict that certain critical topics could be particularly prone to AI-source effects. However, we opted to start with a straightforward and widely applicable, current health topic that was also relevant for our participants.}. 

\subsection{Vaping Prevention as Context to Examine the Source Effects of AI}

The use of e-cigarettes (or vaping) has become a significant public health concern in the last decade, especially because of the high prevalence of e-cigarette use among youth (<18 years of age) and young adults (18-24 years of age). About 20\% of high school and 5\% of middle school students reported vaping in 2020 \citep{wang2021characteristics}; it was also estimated that about 15\% of young adults were using e-cigarettes in 2020 \citep{boakye2022assessment}. Moreover, much of smoking and vaping-related marketing leverages the power of social media - or its capacity in disseminating information and ideas at a rapid speed through networks of people following one another \citep{nahon2013going} - to influence audiences and promote tobacco products \citep{allem2017campaigns, clark2016vaporous, collins2019e-cigarette}. To combat the detrimental effects of vaping, health researchers and professionals have invested significant efforts into developing and testing effective campaign messages \citep{boynton2023perceived, liu2020incorporating, noar2020evaluating, villanti2021identifying}, leading to guidelines for best practices (e.g., Vaping Prevention Resource, 2023). These efforts could be further augmented by the capabilities of LLMs in generating effective health messages \citep{karinshak2023working, lim2023artificial}. 

\subsection{The Current Study and Hypotheses}

The current study examined how human participants respond to vaping prevention messages that were either generated by AI vs. humans by either adding accurate source labels to the messages (source disclosed) or not adding any labels (source not disclosed).

\subsubsection{Effects Perception Ratings as Measure of Evaluation}

Within health campaigns research, one of the most used message evaluation metrics is perceived message effectiveness (PME). According to \cite{baig2019unc}, the PME measure tends to cover two major constructs, message perceptions and effects perception. Message perceptions refer to the extent the messages seem credible and understandable, while effects perception refers to how the message promotes self-efficacy and behavioral intention. \cite{baig2019unc} developed an effects perception scale that focused on examining the extent the message does what it is intended. Existing research showed that effects perception was highly associated with health campaign outcomes such as risk beliefs, attitudes, and behavioral intentions \citep{grummon2022reactions, noar2020evaluating, rohde2021comparison}, meanwhile in some cases message perceptions did not have significant associations with these outcomes. Thus, we used effects perception ratings as people’s measure of the perceived effectiveness of the messages.

Since the influence of source disclosure is a relatively new area of research, to our knowledge, only one study specifically examined how source disclosure would impact people’s ratings of health campaigns messages at the time of writing this manuscript. \cite{karinshak2023working} conducted a set of three exploratory studies that used GPT3 to generate high-quality vaccination promotion messages. The third study, which manipulated source labels, found that prevention messages generated by GPT3 were rated higher in terms of perceived message effectiveness compared to those written by CDC when none of the messages were labeled. However, messages labeled as AI-generated were rated lower in terms of argument strength and perceived message effectiveness compared to those labeled as created by CDC or those not labeled at all. 

Our study had a few aspects that differed from \cite{karinshak2023working} study. For one, our comparison of human-generated messages were tweets, to take into account that much discussion about vaping occurs via social media platforms such as Twitter \citep{lyu2021vaping, wang2023moralization}. Second, we used effects perception measure specifically (rather than the general perceived message effectiveness) as a measure of message evaluation. Still, as existing literature suggests the existence of negative bias against AI-generated content, we posed the following hypothesis:

Hypothesis 1 (H1): People who know the source of the messages will rate AI-generated messages lower and human-generated tweets higher than those who did not know the source.

\subsubsection{Ranking as Measure of Preference}

In addition to effects perception ratings, rankings have also been used in existing research to gather information about preference. Unlike ratings, rankings ask participants to order the messages from the best to the worst, using whatever criteria provided by the researcher and/or determined by the participants \citep{ali2012ordinal}. Rankings have been used extensively in the social sciences to gather data about constructs such as values \citep{abalo2007importance, alwin1985measurement}, and attribute preferences \citep{lagerkvist2013consumer}. Within health communication, ranking measurement was used to examine people’s preferences, including preferred health promotion icons \citep{prasetyo2021evaluation} and factors that influence demand for vaccinations \citep{ozawa2017using}. Though we do not know of any work that examined the influence of source disclosure on people’s ranking of AI-generated vs. human-generated messages, we still predict that the negative bias against AI-generated messages will be exhibited in the ranking of the messages. Thus, we pose the following hypothesis:

H2: Those who know the source will prefer human-generated tweets vs. AI-generated prevention messages.

\subsection{Method}

We pre-registered our hypotheses and procedures at as.predicted.  

\subsubsection{Participants}

A total of 151 young adults (18-24 years of age) were recruited from two study pools and either received course credit (University study pool) or \$2.80 (Prolific; \cite{palan2018prolific}) as compensation for participating in the study. We specifically selected the young adult age group because of the prevalence of vaping in this age demographic \citep{boakye2022assessment}. The local review board approved the study. We discarded the data from nine participants who did not complete the study or who completed the study in under five minutes, leaving 142 participants (\textit{m}$_{age}$ = 20.78, \textit{sd}$_{age}$ = 1.78]; 59\% women) in the final dataset. Power calculations conducted a priori using the WebPower package in R \citep{zhang2018practical} for a mixed ANOVA, with a medium effect size (\textit{f} = 0.25) and significance level $\alpha$ = .05, showed that a total sample size of around 130 (about 65 per group) was enough to detect significant differences between groups at the power level of 0.8. 

\subsubsection{Experimental Messages: Human- and AI-generated }

We relied on previously published procedures to generate messages via a LLM, collect human-generated messages, and select 30 total messages (15 AI, 15 human) for the experiment \citep{lim2023artificial}. For details, see Appendix A. For the sake of relevance and length, we briefly outline the process here. 

To collect human-generated messages, we scraped vaping prevention tweets with hashtags \#dontvape, \#novaping, \#quitvaping, \#stopvaping, \#vapingkills, and \#vapingprevention using the snscrape package \citep{snscrape2021} in Python. After cleaning the tweets, we randomly selected 15 tweets that had been retweeted at least once for the experiment.  

For AI message generation and selection, we generated 500 total vaping prevention messages using the Bloom LLM, and then randomly selected a subset of 15 messages. Bloom is the largest open-source multilingual language model available \citep{scao2022bloom}. As mentioned in previous sections, Bloom, like GPT3, is powered by the transformer neural network, the most advanced ANN system currently available \citep{tunstall2022natural}. Pre-trained with 1.5 TB of pre-processed text from 45 natural and 12 programming languages, Bloom allows for text generation using prompting (inputting the beginning part of the text and the language model completes the text) and a set of statistical parameters. We chose Bloom because of its free cost, full transparency of the training process and training data, and the ability to use it on a local machine via Jupiter notebooks or Google Colab without a special computing system called graphic processing unit (GPU), often required to run large computational tasks. 

\subsubsection{Experimental Procedure and Conditions}

The experiment was conducted online via Qualtrics. Once participants consented to the study, the young adult participants were randomly assigned to one of two groups: control and treatment (\textit{n}$_{control}$ = 72, \textit{n}$_{treatment}$ = 70). Then the survey asked the participants to rate each message on four perceived message effectiveness items and rank the 30 messages (15 AI-generated vs. 15 tweets). The order of the two activities was randomized to control for order effects. The participants in the treatment condition read messages with source labels (e.g., “AI-Generated Message: Nicotine in vapes…”, “Human-Generated Tweet: Nicotine in vapes can…”) while those in the control condition were not provided the source labels. The source labels were true - no deception was used. Upon completing the main experiment, participants completed demographic questions and were debriefed about the study’s purpose.

\subsubsection{Measures}

Study 1 included two main measures. First, we adopted and updated UNC’s perceived message effects, otherwise named effects perceptions (EP), scale \citep{baig2019unc} to fit vaping. The measure included the following four survey items: “This message discourages me from wanting to vape,” “This message makes me concerned about the health effects of vaping,” “This message makes vaping seem unpleasant to me,” and “This message makes vaping seem less appealing to me.” Participants rated each item on a likert scale from 1 (Strongly disagree) to 5 (Strongly agree). Second, for the ranking activity, we asked participants to rank the 30 messages from the best (1) to the worst (30) message by dragging each message to its rank. Finally, the participants answered demographic questions including age. 

\subsubsection{Data Analysis}

All analyses were conducted in R. To examine H1, the responses for the four items of the EP scale were averaged into a composite EP score for each participant; the last item about the appeal of vaping was excluded from the analysis to keep consistent with the results from \cite{baig2019unc}. Then we conducted a mixed ANOVA that examined the influence of source disclosure (disclosed vs. undisclosed) and the message source (AI vs. human) on EP.

For the statistical difference in the mean ranks between the groups, we first subtracted the mean ranks for the human messages from the mean ranks of the AI messages (AI - Human). Thus, if the human-generated messages were on average ranked higher than AI-generated messages, then this difference value would be negative, and vice versa. Using the stats package \citep{chambers1992analysis}, we conducted the Wilcoxon Rank Sum Test, the non-parametric alternative to a two-sample ANOVA. We used the alpha level of $\alpha$ = .05 to test for significance for both mixed ANOVA and Wilcoxon Rank Test. 

In addition, we conducted a supplementary computational analysis. The purpose of this was to extract and compare various textual features of the AI-generated messages and human-generated tweets, showing that the two groups of messages could be adequately compared. The textual methods we used included semantic analysis, n-gram analysis, topic modeling, sentiment analysis, and assessment of readability metrics. These analyses were carried out using Python and R packages including spacy, textacy, vader, topicmodels, and the sentence-transformers \citep{dewilde2020, grun2011topicmodels, honnibal2020, hutto2014vader, reimers2019sentence}. For all computational analysis of tweets, we removed the hashtags used to scrape the tweets. We also removed the prompts from the AI-generated messages for all analyses except semantic analysis. See Appendix B for the results of the supplementary analysis.

\subsubsection{Deviation from Pre-registration}

While the main ideas from the pre-registration remained the same, we altered some of the details of the pre-registration. First, the pre-registration only included the data collection plan for the University sample. We decided to gather additional data from Prolific to make the results more generalizable beyond the University sample and to increase the sample size. Second, we decided to aggregate only the first three out of the four items for the EP measure to be more consistent with the existing literature \citep{baig2019unc}. Finally, for the rank data, we used the Wilcoxon test, which is a two-sample extension of the Kruskal-Wallis test.

\subsection{Results}

First, we present the results from the mixed ANOVA, which tested the influence of source disclosure on message ratings (see Table 1). We find that there was a significant interaction effect between source disclosure and the message source (\textit{F(1,140)} = 4.73, $\eta^2$ = .0018, \textit{p} = .031). As illustrated in Figure 2, this interaction was due to the fact the difference between the AI-generated and human-generated messages was smaller when the source was disclosed compared to when it was not disclosed. This interaction qualified a main effect of message source (\textit{F(1,140)} = 10.25, $\eta^2$ = .0039, \textit{p} = .0017), which indicated overall lower ratings for  human-generated compared to AI-generated messages. Follow-up comparisons conducted separately for each message source (i.e. AI-generated and human-generated messages) revealed  that the EP ratings for AI-generated messages were slightly lower and ratings for human-generated messages were slightly higher when the source was disclosed, yet this difference was not statistically significant (\textit{t(133)} = .82; \textit{p} $>$ .05 for AI-generated messages;  \textit{t(125)} = -.23; \textit{p} $>$ .05 for human-generated messages; see Table 2). Thus, H1 was partially supported.

\begin{table}[hbt!]
\caption{Influence of Source Disclosure on EP Scores}
\label{Table1}
\small
\begin{tabular}{l c c c} 
 \hline
  & \textbf{\textit{F-score}} & \textbf{\textit{p-value}} & \textbf{$\eta^2$} \\ 
 \hline
 SD: Disclosed (vs. Not Disclosed) & .072 & .79 & .00049 \\ 
 MS: Human (vs. AI) & 10.25 & \textbf{.0017} & .0039 	 \\ 
 SD: Disclosed \& MS: Human & 4.73 & \textbf{.031} & .0018	 \\ 
 \hline
\end{tabular}
{\raggedright \textit{Note. SD = Experimental Group; MS = Message Source.} \par}
\end{table}

\begin{figure}[hbt!]
	\includegraphics[width=0.5\textwidth]{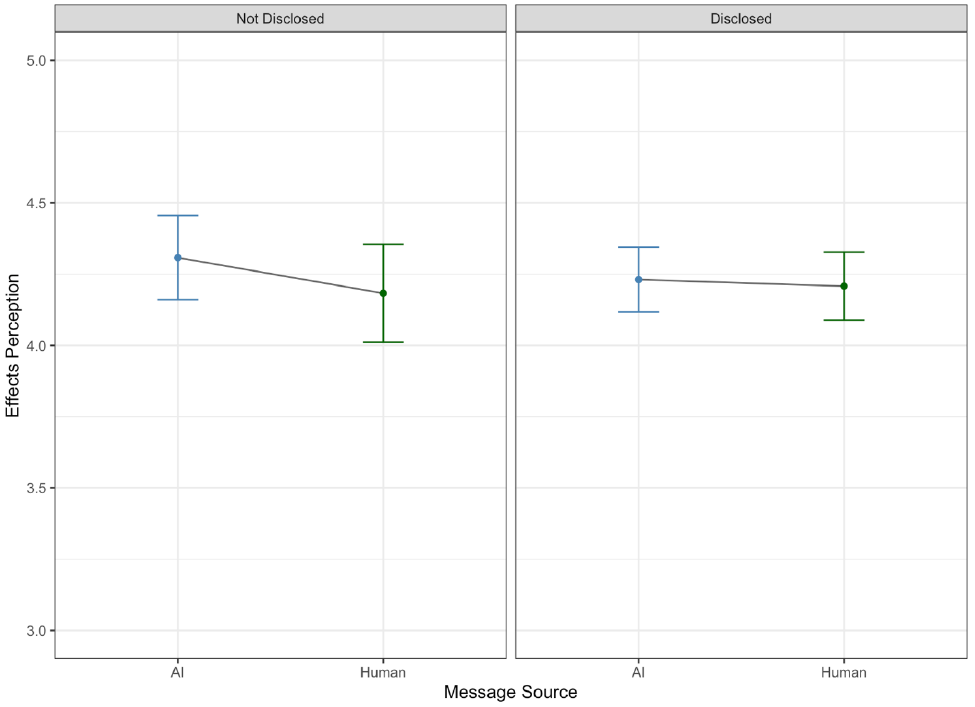}	
	\caption{Mean Effects Perception Scores by Experimental Condition.} 
	\label{fig_mom1}%
\end{figure}

\begin{table}[hbt!]
\caption{The Effect of Source Disclosure by Message Source}
\label{Table2}
\scriptsize
\begin{tabular}{l c c c} 
 \hline
  & \multicolumn{2}{c}{Mean (Standard Deviation)} & t-score \textit{(p-value)} \\ 
 \hline
 & Source Not Disclosed & Source Disclosed & \\ 
 \hline
 AI-Generated & 4.31 (.63) & 4.23 (.48) & .82 \textit{(.41)} \\ 
 Human-Generated & 4.18 (.73) & 4.21 (.50) & -.23 \textit{(.82)}	 \\ 
 \hline
\end{tabular}
\end{table}

To test H2, we compared the difference in the mean ranks of AI and human-generated messages (AI mean rank - Human mean rank; see Figure 3) using the Wilcoxon Sum Rank Test.  For the rank activity, the lower quantitative value represented a higher relative quality rank, with 1 representing the best message. Thus, the smaller differences in rank suggested a lower quantitative value for AI mean rank, hence a higher preference for AI-generated messages. We found that the median difference in rank for participants who knew the source, \textit{Mdn} = -.6, was slightly higher than the median difference in rank for participants who did not know the source, \textit{Mdn}  = -.87, though this difference was not statistically significant (\textit{W} = 2652.5, \textit{p} $>$ .05).

\begin{figure}[hbt!]
	\includegraphics[width=0.5\textwidth]{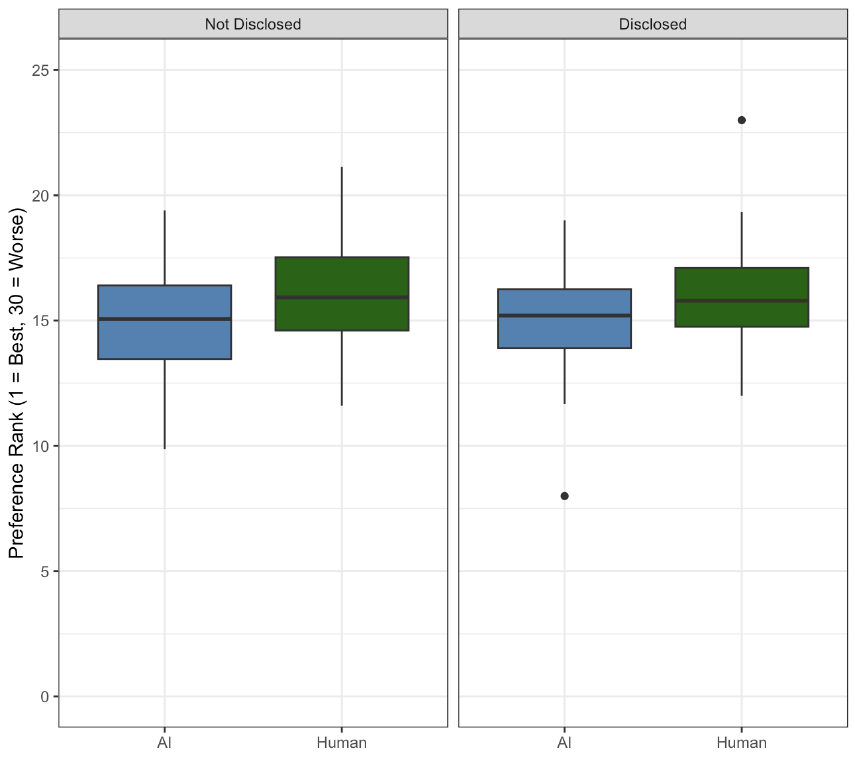}	
	\caption{Mean Rank Difference Scores by Experimental Condition.} 
	\label{fig_mom2}%
\end{figure}

\subsection{Study 1 Discussion}

Study 1 examined how disclosing the source of a message as coming from an AI (vs. humans) influenced the evaluations of the messages and the preferences for the message source. Our H1 was partially supported – source disclosure significantly decreased the ratings difference between AI and human-generated messages. However, follow-up mean comparisons by message source showed that ratings stayed statistical consistent between non source disclosure and source disclosure conditions. This finding is generally aligned with findings from \cite{karinshak2023working}. However, our H2, which addressed the ranking task that required participants to make an active selection to express their preferences about messages, was not supported. This could have occurred for many reasons, one of which is that ranking all 30 messages may have required too much cognitive effort. To further inspect source effects of AI-generated messages, we conducted a follow-up study (Study 2), examining individual differences that could boost or buffer the effects of source disclosure. For instance, participants could vary in their attitudes about the use of and general sentiment towards AI, which in turn could influence their judgments of AI-generated content. Thus, we examined attitudes towards AI as a potential factor in Study 2.

\section{Study 2}

Study 2 replicated the source disclosure manipulation from Study 1 with a few modifications. First, we assessed people’s preference for messages via message selection (top 5 out of 30) rather than the ranking task to decrease the participants’ cognitive burden of comparing all 30 messages. Next, we examined how the influence of source disclosure on the evaluation and selection of AI-generated messages varied by the level of negative attitudes towards AI. 

\subsection{Negative Attitudes Towards AI as Moderators}

\cite{schepman2023general} created a scale about general attitudes toward AI (GAAIS). A major part of the measure is based on the concept of trust in the capabilities and the uses of AI. The paper showed that GAAIS was associated with psychological features such as the Big Five personality, showing that it can be used to represent various individual differences that could exist when processing messages generated by AI. For example, \cite{bellaiche2023humans} examined the association between attitudes towards AI and people’s judgments of art labeled as AI-created or human-created. In this study, we adopted the negative attitudes towards AI subscale, which included people’s concerns about and negative sentiment towards AI, as a moderator. Adopting the negative attitudes towards AI subscale of GAAIS, we posited the following hypotheses:

H3: Negative attitude toward AI will moderate the influence of source disclosure on the evaluation of prevention messages.

H4: Negative attitude toward AI will moderate the influence of source disclosure on the preference for AI as the message source.

\subsection{Method}

\subsubsection{Participants}

As with study 1, we used two platforms to recruit participants, one administered by the university and the other by Prolific. A total of 216 adults recruited from the study pools either received course credit (University study pool) or \$2.80 (Prolific; \cite{palan2018prolific}) as compensation for participating in the study. To generalize the findings of Study 1 beyond young adults, we extended the participant pool for the Prolific platform to all adults . The local review board approved the study. We discarded the data from 33 participants who did not complete the study, completed the study in under five minutes, and failed to pass the manipulation check questions, leaving 183 participants (\textit{m}$_{age}$ = 33.83, \textit{sd}$_{age}$ = 14.42; 56\% women) in the final dataset. 

\subsubsection{Experimental Procedure, Measures, and Data Analysis}

The same 30 messages from Study 1 were tested in the main study. The experiment followed the same procedure as study 1 (\textit{n}$_{control}$ = 94, \textit{n}$_{treatment}$ = 89) with the following modification: instead of ranking the messages, we asked participants to select the 5 best messages from the pool instead of having them rank all messages. This was done because, in campaign practice, the best-in-show messages are chosen from a larger pool of candidates. Moreover, having participants and all messages is rather taxing and we expected better compliance with a more focused task. Upon completing the main experiment, participants answered background and demographics questions that included questions about their attitudes towards AI.

The negative attitude toward AI scale asked people to rate 8 items related to negative attitudes (e.g., “I shiver with discomfort when I think about future uses of Artificial Intelligence”) from a scale of 1 (strongly disagree) to 5 (strongly agree) \citep{schepman2023general}. The overall mean was 3.04, with a standard deviation of .80. The demographics questions stayed the same as in study 1.

All analyses were conducted in R. First, we calculated the average score for negative attitudes towards AI. To examine how the influence of source disclosure on EP of AI vs. human-generated messages differed by the extent of negative attitude (H3), we fitted a mixed effects linear regression model. The models allowed for the intercept to vary by participant, to take into consideration of the repeated measures design. To examine how the effect of source disclosure on source preference differed by negative attitude (H4), we first calculated how many AI-generated messages were selected (out of 3), and then fitted a Poisson regression model. 

\subsection{Results}

For the EP ratings, we found that there was a significant three-way interaction of source disclosure, message source (AI vs. Human), and the extent of having a negative attitude toward AI (\textit{b} = -.14, \textit{SE} = .047, \textit{p} = .0029; see Table 3). In other words, the influence of source disclosure on the evaluation of the AI-generated messages vs. human-generated messages differed by the level of negative attitudes towards AI. A deeper inspection of the moderation effect shows that for both AI-generated and human-generated messages, source disclosure led to slightly higher EP ratings among participants with lower levels of negative attitudes towards AI, whereas it led to slightly lower EP ratings among those with higher levels of negative attitudes towards AI (see Table 4). Interestingly, when the source was disclosed, the more negative attitudes the participants had towards AI, the higher they rated the AI-generated messages, whereas the ratings of human-generated messages generally stayed flat (see Figure 4). 

\begin{table}[hbt!]
\caption{Results from Mixed Effects Linear Model}
\label{Table3}
\scriptsize
\begin{tabular}{l c c c c} 
 \hline
  \textbf{Term} & \textbf{\textit{Estimate}} &  \textbf{\textit{S.E.}} &  \textbf{\textit{t-score}} & \textbf{\textit{p-value}} \\ 
 \hline
 Intercept & 3.32 & .30 & 10.98 & \textbf{$<$.001} \\ 
 SD: Disclosed (vs. Not Disclosed) & .61 & .45 & 1.34 & .18 \\ 
 MS: Human (vs. AI) & -.46 & .098 & -4.69 & \textbf{$<$.001} \\ 
 Negative Attitude Towards AI & .25 & .097 & 2.61 & \textbf{.0098} \\ 
 MS: Human \& SD: Disclosed & .48 & .15 & 3.26 & \textbf{.0011} \\ 
 SD: Disclosed \& Negative Attitude & -.20 & .14 & -1.36 & .18 \\ 
 MS: Human \& Negative Attitude & .099 & .031 & 3.15 & \textbf{.0017} \\ 
 MS: Human \& SD: Disclosed \& Neg Attitude & -.14 & .047 & -2.98 & \textbf{.0029} \\ 
 \hline
\end{tabular}
{\raggedright \textit{Note. SD = Experimental Group; MS = Message Source; S.E. = Standard Error} \par}
\end{table}

\begin{table}[hbt!]
\caption{Pairwise Comparison of EP by Negative Attitudes Towards AI}
\label{Table4}
\scriptsize
\begin{tabular}{l c c c c} 
 \hline
   & \makecell{\textbf{\textit{Disclosed -}} \\ \textbf{\textit{NonDisclosed}}} &  \textbf{\textit{S.E.}} &  \textbf{\textit{z-score}} & \textbf{\textit{p-value}} \\ 
 \hline
 AI-Generated Messages & & & & \\ 
 \indent\hspace{.3cm}\textit{Negative Attitudes Towards AI = 2.24$*$} & .17 & .16 & 1.04 & .30 \\ 
 \indent\hspace{.3cm}\textit{Negative Attitudes Towards AI = 3.04} & .013 & .12 & .11 & .91 \\ 
 \indent\hspace{.3cm}\textit{Negative Attitudes Towards AI = 3.84} & -.14 & .16 & -.89 & .37 \\ 
 Human-Generated Messages & & & & \\ 
 \indent\hspace{.3cm}\textit{Negative Attitudes Towards AI = 2.24$*$} & .34 & .16 & 2.06 & \textbf{.04} \\ 
 \indent\hspace{.3cm}\textit{Negative Attitudes Towards AI = 3.04} & .068 & .12 & .59 & .56 \\ 
 \indent\hspace{.3cm}\textit{Negative Attitudes Towards AI = 3.84} & -.20 & .16 & -1.23 & .22 \\ 
 \hline
\end{tabular}
{\raggedright \textit{Note. S.E. = Standard Error; *3.04 = overall mean; 2.24 = mean - 1 standard deviation; 3.84 = mean + 1 standard deviation} \par}
\end{table}

\begin{figure}
	\includegraphics[width=0.5\textwidth]{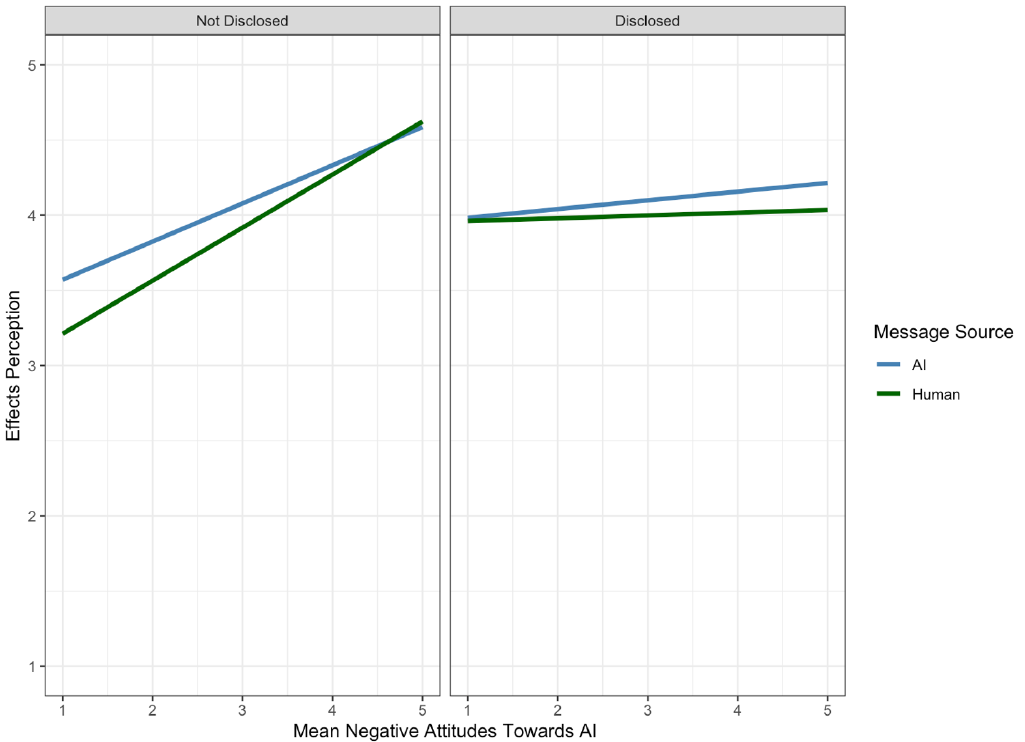}	
	\caption{Predicted EP Ratings with Negative Attitudes as Moderator} 
	\label{fig_mom3}%
\end{figure}

\newpage

Table 5 shows the results for messages selection. There was no moderation effect of negative attitudes toward AI (\textit{b} = -.042, \textit{SE} = .12, \textit{p} $>$ .05), and H4 was not supported. A deeper inspection of the results showed that those who knew the source were likely to select less number of AI-generated messages compared to those who did not know the source for those with moderate level of negative attitudes towards AI (see Figure 5 and Table 6). 

\begin{table}[hbt!]
\caption{Attitudes Towards AI and Selection of AI-Generated Messages}
\label{Table5}
\scriptsize
\begin{tabular}{l c c c c} 
 \hline
   & \textbf{\textit{b}} &  \textbf{\textit{S.E.}} &  \textbf{\textit{z-score}} & \textbf{\textit{p-value}} \\ 
 \hline
 Intercept & .74 & .25 & 2.97 & \textbf{.003} \\ 
 SD: Disclosed (vs. Not Disclosed) & -.09 & .39 & -.24 & .81 \\ 
 Negative Attitudes Towards AI & .07 & .079 & .90 & .37 \\ 
 SD: Disclosed \& Negative Attitudes Towards AI & -.04 & .12 & -.34 & .73 \\ 
 \hline
\end{tabular}
{\raggedright \textit{Note. S.E. = Standard Error} \par}
\end{table}

\begin{table}[hbt!]
\caption{Pairwise Comparison of AI-Message Selection by Negative Attitudes Towards AI}
\label{Table6}
\scriptsize
\begin{tabular}{l c c c c} 
 \hline
    &  \makecell{\textbf{\textit{Disclosed -}} \\ \textbf{\textit{NonDisclosed}}} &
    \textbf{\textit{S.E.}} &
    \textbf{\textit{z-score}} & \textbf{\textit{p-value}} \\ 
 \hline
 Negative Attitudes Towards AI = 2.24$*$ & -.42 & .31 & -1.33 & .18 \\ 
 Negative Attitudes Towards AI = 3.04 & -.51 & .23 & -2.27 & \textbf{.023} \\ 
 Negative Attitudes Towards AI = 3.84 & -.62 & .33 & -1.88 & .06 \\ 
 \hline
\end{tabular}
{\raggedright \textit{Note. S.E. = Standard Error; *3.04 = overall mean; 2.24 = mean - 1 standard deviation; 3.84 = mean + 1 standard deviation} \par}
\end{table}

\begin{figure}[hbt!]
	\includegraphics[width=0.5\textwidth]{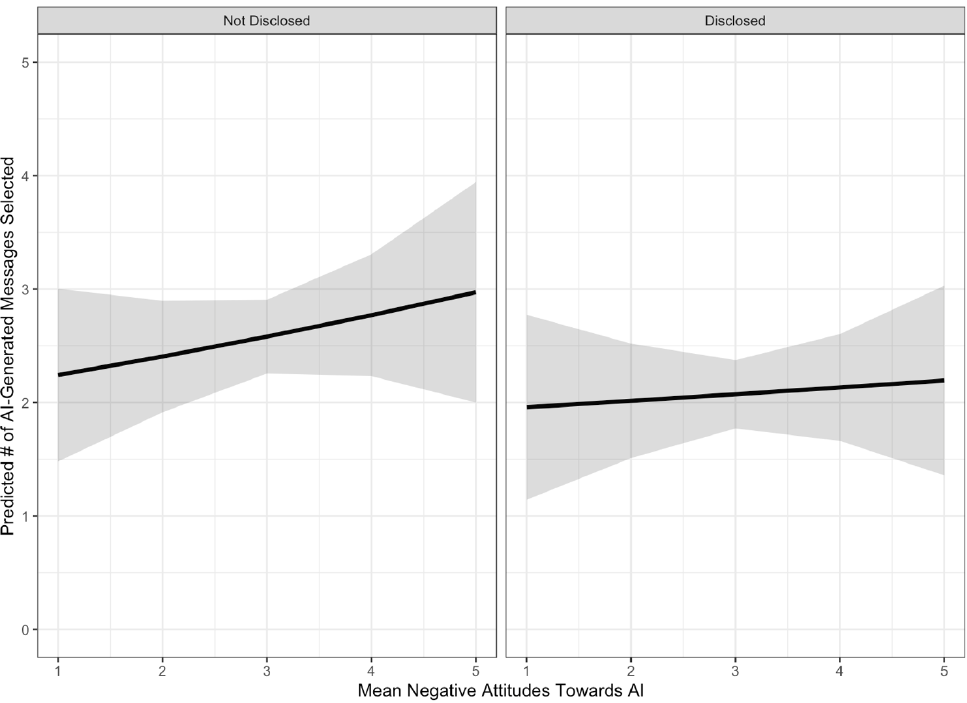}	
	\caption{Predicted Number of AI-Generated Messages Selected} 
	\label{fig_mom4}%
\end{figure}

\subsection{Study 2 Discussion}

Study 2 examined whether negative attitudes towards AI moderated the influence of source disclosure on the evaluation of and preference for AI-generated vs. human-generated messages. For EP ratings, having a negative attitude toward AI emerged as a significant moderator, supporting H3. Specifically, at lower levels of negative attitudes towards AI, the ratings for AI-generated and human-generated messages were slightly higher when the source was disclosed vs. not disclosed (albeit not statistically significant for AI-generated messages), whereas the opposite was observed at higher levels of negative attitudes towards AI (not statistically significant). This result suggests the existence of a slight bias against AI-generated messages.

However, for the participants who knew the source, the EP ratings for AI-generated messages compared to those for human-generated messages increased with the level of negative attitudes towards AI. While deeper inspection is needed to fully unpack this phenomenon, one explanation could be “source involvement”. In other words, the level of negative attitudes towards AI could have determined how closely they examined the messages: those with greater levels of negative attitudes towards AI could have paid closer attention to the content of the messages compared to those with less negative attitudes towards AI. Another explanation is that the negative attitudes towards AI measure could be a bit too general. In the case of AI message generation, people are heavily involved in the process. In the case of AI message generation, people are heavily involved in the process (see Appendix A for details of how the messages used for this study were crafted). Thus, it is possible that people’s general attitudes towards AI did not play as large of a role as we expected in their evaluation of the generated messages. 

For message selection, we did not find any significant moderating effects; thus, our H4 was not supported. However, source disclosure significantly decreased the number of AI-generated messages selected for those with moderate levels of negative attitudes towards AI. These results further provide support for people’s preference against AI-generated messages. We discuss the theoretical and practical implications of our findings in the next section.

\section{Overall Discussion}

\subsection{Summary of the Findings}

Overall, our two studies provide qualified support for our hypotheses. We found that disclosing the source led to lower ratings of AI-generated messages (partially supporting H1 in Study 1) and that negative attitudes toward AI moderated this effect (supporting H3 in Study 2).  Though the analyses of ranking and message selection tasks did not support our hypotheses (H2 in Study 1 and H4 in Study 2), they revealed interesting effects: source disclosure decreased the number of AI-generated messages selected for those with moderate levels of negative attitudes towards AI. These results suggest a slight negative bias against AI-generated messages, aligning with previous studies that showed hesitation and slight negative bias against the communicative content when participants believed AI was involved in the process \citep{asscher2023human, jakesch2019ai-mediated, karinshak2023working, liu2022will, ragot2020ai, shank2023ai}. 

\subsection{Implications for Source Effects Research}

This paper contributes to the emerging area of study at the intersection of communication and AI by being one of the first papers to examine how knowing the source changes people’s evaluation of and preference for AI-generated messages. 

The source of a message has always been an integral part of theories and models of communication and persuasion, even going back to Aristotle’s rhetoric theory \citep{murphy1981rhetoric}. Likewise, early models of social scientific communication research, such as Berlo’s SMCR model \citep{berlo1960process}, Lasswell’s model of communication \citep{lasswell1948structure}, and even the Shannon-Weaver model of communication \citep{shannon1948mathematical} all included components about the source, or the creator and deliverer of the message. Since then, a plethora of studies have studied source effects, or how various characteristics of the source impact the way people receive, process, and subsequently make judgments about the message. These studies often manipulated certain aspects about the source (e.g., expert vs. nonexpert; \cite{clark2012source}) and examined in which scenarios the various levels led to greater persuasive outcomes (e.g., when people had little information about a product, they relied on expert sources, but not necessarily when they had more information; \cite{ratneshwar1991comprehension}). 

With the rise of AI-based technologies such as LLMs, source effects have once again come to the forefront of communication research, but the notion of “source” for AI-generated messages is quite complex. In particular, the message generation process for LLMs generally consists of the following steps: First, a human user feeds prompts, or intentionally crafted instructions or beginning parts of the message, to the LLM; second, the user adjusts the parameters, such as how many messages should be crafted and other factors; third, the LLM generates the messages according to step 1 and 2. Thus, in this process, it is actually a human who initiates the message creation sequence, whereas the LLM only completes the message generation command. It seems plausible to assume that people’s knowledge of AI (and their perceptions of its expertise, trustworthiness, etc.), will impact their evaluations. For example, we may surmise that it would not only matter that a message was AI-generated, but also whom people believe to have started the process. In other words, if people think that the AI message generation was initiated by expert organizations, such as the Center for Disease Control (CDC), evaluation might differ compared to AI-generated messages initiated by general users of social media platforms, or even by agents from a foreign country. In sum, with AI, there is an intersection of source effects, perceptions of AI, and various social-cognitive inferences about creator, intent, and expertise. Going forward, it will thus be important to comprehensively study these topics. Based on the present results, we can say that there are small but significant effects of source disclosure, consistent with a small preferential treatment for human-generated messages. 

\subsection{Implications for Public Health Campaigns}

The current results have interesting implications for research on health message generation and dissemination. With the advent of AI-language models, it has become extremely easy to generate high-quality health messages about any given topic. This potential can either be a blessing or a curse, depending on the source and their intent. For instance, if the CDC leveraged the power of AI for health message generation, this would be seen as largely beneficial; however, malicious actors could also leverage AI to spread fake news - or even just promote unhealthy products (e.g., cigarettes). Indeed, there are already commercial applications of AI-LLMs for copywriting purposes, and these could also be used to influence users towards unhealthy, risky, or other kinds of behaviors. Thus, more work is needed to explore how these aspects intersect with the topic of AI-as-message-source as well as the influence of source disclosure.

A related concern is about the factual truthfulness of health-related claims. It is well known that although LLMs are capable of generating persuasive messages, they are prone to hallucinations \citep{kaddour2023challenges, zhang2023siren}. Although the creators of AI systems are investing large efforts to minimize such false generations, this is still an unsolved problem of the underlying technology, which will affect the evaluation of AI systems \citep{marcus2018deep, marcus2020next}, particularly whether AIs are seen as knowledgeable, reliable, and trustworthy. In sum, while we can expect that AI-generated messages will increasingly find their way into real-world health campaigns, numerous questions persist about their accuracy and the intent of the humans generating the messages using AI. At this point in time, the dynamically evolving landscape of AI-language generation systems prevents any final answers to these questions. Rather, longitudinal research would be needed to assess how people think about AI sources, how they adapt to the increasing prevalence of AI content, and how their evaluations are influenced by contextual factors. 

\subsection{Limitations, Future Avenues, and Ethical Considerations}

As with all research, several limitations that require future research and important ethical considerations are worth highlighting. One limitation is that this study used tweets as messages. It would be interesting to examine other kinds of health messages, such as longer flyers and posters. The decision to use tweets was made because we wanted to take into account user-generated messages and because tweets have become a rather widespread form of health communication content that also gets used by the CDC and other key health organizations. In addition, the topic of AI-generated health messages raises ethical questions. In particular, the regulatory framework around these topics is currently in flux, and discussions about mandatory labeling of AI-generated content have barely even begun. Furthermore, the allowed use cases for AI content generation are also debated. For instance, using AI to generate medical diagnoses is explicitly prohibited by the creators, but generating general health information falls within the range of acceptable use \citep{bigscience2022}.  

\section{Summary and Conclusion}

Taken together, we examined the influence of source disclosure on evaluations of AI-generated messages. We found that source disclosure (i.e., labeling the source of a message as AI vs. human) significantly impacted the evaluation of the messages, albeit the effects were of relatively small magnitude, but did not significantly alter message rankings. Moreover, in study 2 we found a significant moderating effect of negative attitudes toward AI on message evaluation. Our results show that at the point when we conducted our research, humans appear to exhibit a small preference for human-generated content if they know the source, but AI-generated messages are evaluated as equally good, if not better, if the source stays unknown. These results highlight the role of source factors for communication, and they have implications for the potential labeling of AI-generated content in the context of health promotion efforts.

\section*{Funding Statement}

This work was supported in part through Michigan State University’s Institute for Cyber-Enabled Research Cloud Computing Fellowship, with computational resources and services provided by Information Technology Services and the Office of Research and Innovation at Michigan State University. The work was additionally supported by the Strosacker Grant from Michigan State University’s Health and Risk Communication Center. 

\bibliographystyle{elsarticle-harv}
\bibliography{main}

\begin{thebibliography}{73}
\expandafter\ifx\csname natexlab\endcsname\relax\def\natexlab#1{#1}\fi
\providecommand{\url}[1]{\texttt{#1}}
\providecommand{\href}[2]{#2}
\providecommand{\path}[1]{#1}
\providecommand{\DOIprefix}{doi:}
\providecommand{\ArXivprefix}{arXiv:}
\providecommand{\URLprefix}{URL: }
\providecommand{\Pubmedprefix}{pmid:}
\providecommand{\doi}[1]{\href{http://dx.doi.org/#1}{\path{#1}}}
\providecommand{\Pubmed}[1]{\href{pmid:#1}{\path{#1}}}
\providecommand{\bibinfo}[2]{#2}
\ifx\xfnm\relax \def\xfnm[#1]{\unskip,\space#1}\fi
%Type = Article
\bibitem[{Abalo et~al.(2007)Abalo, Varela and Manzano}]{abalo2007importance}
\bibinfo{author}{Abalo, J.}, \bibinfo{author}{Varela, J.}, \bibinfo{author}{Manzano, V.}, \bibinfo{year}{2007}.
\newblock \bibinfo{title}{Importance values for importance--performance analysis: A formula for spreading out values derived from preference rankings}.
\newblock \bibinfo{journal}{Journal of Business Research} \bibinfo{volume}{60}, \bibinfo{pages}{115--121}.
%Type = Article
\bibitem[{Ali and Ronaldson(2012)}]{ali2012ordinal}
\bibinfo{author}{Ali, S.}, \bibinfo{author}{Ronaldson, S.}, \bibinfo{year}{2012}.
\newblock \bibinfo{title}{Ordinal preference elicitation methods in health economics and health services research: using discrete choice experiments and ranking methods}.
\newblock \bibinfo{journal}{British Medical Bulletin} \bibinfo{volume}{103}, \bibinfo{pages}{21--44}.
%Type = Article
\bibitem[{Allem et~al.(2017)Allem, Escobedo, Chu, Soto, Cruz and Unger}]{allem2017campaigns}
\bibinfo{author}{Allem, J.}, \bibinfo{author}{Escobedo, P.}, \bibinfo{author}{Chu, K.}, \bibinfo{author}{Soto, D.}, \bibinfo{author}{Cruz, T.}, \bibinfo{author}{Unger, J.}, \bibinfo{year}{2017}.
\newblock \bibinfo{title}{Campaigns and counter campaigns: reactions on twitter to e-cigarette education}.
\newblock \bibinfo{journal}{Tobacco Control} \bibinfo{volume}{26}, \bibinfo{pages}{226--229}.
%Type = Article
\bibitem[{Alwin and Krosnick(1985)}]{alwin1985measurement}
\bibinfo{author}{Alwin, D.}, \bibinfo{author}{Krosnick, J.}, \bibinfo{year}{1985}.
\newblock \bibinfo{title}{The measurement of values in surveys: A comparison of ratings and rankings}.
\newblock \bibinfo{journal}{Public Opinion Quarterly} \bibinfo{volume}{49}, \bibinfo{pages}{535--552}.
%Type = Article
\bibitem[{Asscher and Glikson(2023)}]{asscher2023human}
\bibinfo{author}{Asscher, O.}, \bibinfo{author}{Glikson, E.}, \bibinfo{year}{2023}.
\newblock \bibinfo{title}{Human evaluations of machine translation in an ethically charged situation}.
\newblock \bibinfo{journal}{New Media \& Society} \bibinfo{volume}{25}, \bibinfo{pages}{1087--1107}.
%Type = Article
\bibitem[{Baig et~al.(2019)Baig, Noar, Gottfredson, Boynton, Ribisl and Brewer}]{baig2019unc}
\bibinfo{author}{Baig, S.}, \bibinfo{author}{Noar, S.}, \bibinfo{author}{Gottfredson, N.}, \bibinfo{author}{Boynton, M.}, \bibinfo{author}{Ribisl, K.}, \bibinfo{author}{Brewer, N.}, \bibinfo{year}{2019}.
\newblock \bibinfo{title}{Unc perceived message effectiveness: validation of a brief scale}.
\newblock \bibinfo{journal}{Annals of Behavioral Medicine} \bibinfo{volume}{53}, \bibinfo{pages}{732--742}.
%Type = Article
\bibitem[{Bellaiche et~al.(2023)Bellaiche, Shahi, Turpin, Ragnhildstveit, Sprockett, Barr, Christensen and Seli}]{bellaiche2023humans}
\bibinfo{author}{Bellaiche, L.}, \bibinfo{author}{Shahi, R.}, \bibinfo{author}{Turpin, M.}, \bibinfo{author}{Ragnhildstveit, A.}, \bibinfo{author}{Sprockett, S.}, \bibinfo{author}{Barr, N.}, \bibinfo{author}{Christensen, A.}, \bibinfo{author}{Seli, P.}, \bibinfo{year}{2023}.
\newblock \bibinfo{title}{Humans versus ai: whether and why we prefer human-created compared to ai-created artwork}.
\newblock \bibinfo{journal}{Cognitive Research} \bibinfo{volume}{8}, \bibinfo{pages}{1--22}.
%Type = Book
\bibitem[{Berlo(1960)}]{berlo1960process}
\bibinfo{author}{Berlo, D.}, \bibinfo{year}{1960}.
\newblock \bibinfo{title}{The Process of Communication}.
\newblock \bibinfo{publisher}{Holt, Rinehart, and Winston}.
%Type = Misc
\bibitem[{Bigscience(2022)}]{bigscience2022}
\bibinfo{author}{Bigscience}, \bibinfo{year}{2022}.
\newblock \bibinfo{title}{Bigscience rail license v1.0}.
\newblock \bibinfo{howpublished}{https://huggingface.co/spaces/bigscience/license}.
%Type = Article
\bibitem[{Boakye et~al.(2022)Boakye, Osuji, Erhabor, Obisesan, Osei, Mirbolouk and Blaha}]{boakye2022assessment}
\bibinfo{author}{Boakye, E.}, \bibinfo{author}{Osuji, N.}, \bibinfo{author}{Erhabor, J.}, \bibinfo{author}{Obisesan, O.}, \bibinfo{author}{Osei, A.}, \bibinfo{author}{Mirbolouk, M.}, \bibinfo{author}{Blaha, M.}, \bibinfo{year}{2022}.
\newblock \bibinfo{title}{Assessment of patterns in e-cigarette use among adults in the us, 2017-2020}.
\newblock \bibinfo{journal}{JAMA Network Open} \bibinfo{volume}{5}, \bibinfo{pages}{e2223266--e2223266}.
%Type = Article
\bibitem[{Boynton et~al.(2023)Boynton, Sanzo, Brothers, Kresovich, Sutfin, Sheeran and Noar}]{boynton2023perceived}
\bibinfo{author}{Boynton, M.}, \bibinfo{author}{Sanzo, N.}, \bibinfo{author}{Brothers, W.}, \bibinfo{author}{Kresovich, A.}, \bibinfo{author}{Sutfin, E.}, \bibinfo{author}{Sheeran, P.}, \bibinfo{author}{Noar, S.}, \bibinfo{year}{2023}.
\newblock \bibinfo{title}{Perceived effectiveness of objective elements of vaping prevention messages among adolescents}.
\newblock \bibinfo{journal}{Tobacco Control} \bibinfo{volume}{32}, \bibinfo{pages}{e228--e235}.
%Type = Article
\bibitem[{Bubeck et~al.(2023)Bubeck, Chandrasekaran, Eldan, Gehrke, Horvitz, Kamar and Zhang}]{bubeck2023sparks}
\bibinfo{author}{Bubeck, S.}, \bibinfo{author}{Chandrasekaran, V.}, \bibinfo{author}{Eldan, R.}, \bibinfo{author}{Gehrke, J.}, \bibinfo{author}{Horvitz, E.}, \bibinfo{author}{Kamar, E.}, \bibinfo{author}{Zhang, Y.}, \bibinfo{year}{2023}.
\newblock \bibinfo{title}{Sparks of artificial general intelligence: Early experiments with gpt-4}.
\newblock \bibinfo{journal}{arXiv preprint arXiv:2303.12712} .
%Type = Article
\bibitem[{Castelo and Ward(2021)}]{castelo2021conservatism}
\bibinfo{author}{Castelo, N.}, \bibinfo{author}{Ward, A.}, \bibinfo{year}{2021}.
\newblock \bibinfo{title}{Conservatism predicts aversion to consequential artificial intelligence}.
\newblock \bibinfo{journal}{Plos One} \bibinfo{volume}{16}, \bibinfo{pages}{e0261467}.
%Type = Incollection
\bibitem[{Chambers et~al.(1992)Chambers, Freeny and Heiberger}]{chambers1992analysis}
\bibinfo{author}{Chambers, J.}, \bibinfo{author}{Freeny, A.}, \bibinfo{author}{Heiberger, R.}, \bibinfo{year}{1992}.
\newblock \bibinfo{title}{Analysis of variance; designed experiments}, in: \bibinfo{booktitle}{Statistical Models in S}. \bibinfo{publisher}{Routledge}, pp. \bibinfo{pages}{145--193}.
%Type = Incollection
\bibitem[{Chen and Chaiken(1999)}]{chen1999heuristic}
\bibinfo{author}{Chen, S.}, \bibinfo{author}{Chaiken, S.}, \bibinfo{year}{1999}.
\newblock \bibinfo{title}{The heuristic-systematic model in its broader context}, in: \bibinfo{booktitle}{Dual-process theories in social psychology}. \bibinfo{publisher}{The Guilford Press}, pp. \bibinfo{pages}{73--96}.
%Type = Article
\bibitem[{Clark et~al.(2016)Clark, Jones, Williams, Kurti, Norotsky, Danforth and Dodds}]{clark2016vaporous}
\bibinfo{author}{Clark, E.}, \bibinfo{author}{Jones, C.}, \bibinfo{author}{Williams, J.}, \bibinfo{author}{Kurti, A.}, \bibinfo{author}{Norotsky, M.}, \bibinfo{author}{Danforth, C.}, \bibinfo{author}{Dodds, P.}, \bibinfo{year}{2016}.
\newblock \bibinfo{title}{Vaporous marketing: Uncovering pervasive electronic cigarette advertisements on twitter}.
\newblock \bibinfo{journal}{PloS One} \bibinfo{volume}{11}, \bibinfo{pages}{e0157304}.
%Type = Article
\bibitem[{Clark et~al.(2012)Clark, Wegener, Habashi and Evans}]{clark2012source}
\bibinfo{author}{Clark, J.}, \bibinfo{author}{Wegener, D.}, \bibinfo{author}{Habashi, M.}, \bibinfo{author}{Evans, A.}, \bibinfo{year}{2012}.
\newblock \bibinfo{title}{Source expertise and persuasion: The effects of perceived opposition or support on message scrutiny}.
\newblock \bibinfo{journal}{Personality and Social Psychology Bulletin} \bibinfo{volume}{38}, \bibinfo{pages}{90--100}.
%Type = Article
\bibitem[{Collins et~al.(2019)Collins, Glasser, Abudayyeh, Pearson and Villanti}]{collins2019e-cigarette}
\bibinfo{author}{Collins, L.}, \bibinfo{author}{Glasser, A.}, \bibinfo{author}{Abudayyeh, H.}, \bibinfo{author}{Pearson, J.}, \bibinfo{author}{Villanti, A.}, \bibinfo{year}{2019}.
\newblock \bibinfo{title}{E-cigarette marketing and communication: How e-cigarette companies market e-cigarettes and the public engages with e-cigarette information}.
\newblock \bibinfo{journal}{Nicotine and Tobacco Research} \bibinfo{volume}{21}, \bibinfo{pages}{14--24}.
%Type = Misc
\bibitem[{DeWilde(2020)}]{dewilde2020}
\bibinfo{author}{DeWilde, B.}, \bibinfo{year}{2020}.
\newblock \bibinfo{title}{Textacy: Nlp, before and after spacy}.
\newblock \bibinfo{howpublished}{https://github.com/chartbeat-labs/textacy}.
\newblock \bibinfo{note}{Accessed September 10, 2022}.
%Type = Article
\bibitem[{von Eschenbach(2021)}]{von2021transparency}
\bibinfo{author}{von Eschenbach, W.J.}, \bibinfo{year}{2021}.
\newblock \bibinfo{title}{Transparency and the black box problem: Why we do not trust ai}.
\newblock \bibinfo{journal}{Philosophy \& Technology} \bibinfo{volume}{34}, \bibinfo{pages}{1607--1622}.
%Type = Article
\bibitem[{Grummon et~al.(2022)Grummon, Hall, Mitchell, Pulido, Sheldon, Noar, Ribisl and Brewer}]{grummon2022reactions}
\bibinfo{author}{Grummon, A.}, \bibinfo{author}{Hall, M.}, \bibinfo{author}{Mitchell, C.}, \bibinfo{author}{Pulido, M.}, \bibinfo{author}{Sheldon, J.}, \bibinfo{author}{Noar, S.}, \bibinfo{author}{Ribisl, K.}, \bibinfo{author}{Brewer, N.}, \bibinfo{year}{2022}.
\newblock \bibinfo{title}{Reactions to messages about smoking, vaping and covid-19: Two national experiments}.
\newblock \bibinfo{journal}{Tobacco Control} \bibinfo{volume}{31}, \bibinfo{pages}{402--410}.
%Type = Article
\bibitem[{Gr{\"u}n and Hornik(2011)}]{grun2011topicmodels}
\bibinfo{author}{Gr{\"u}n, B.}, \bibinfo{author}{Hornik, K.}, \bibinfo{year}{2011}.
\newblock \bibinfo{title}{topicmodels: An r package for fitting topic models}.
\newblock \bibinfo{journal}{Journal of Statistical Software} \bibinfo{volume}{40}, \bibinfo{pages}{1--30}.
%Type = Article
\bibitem[{Hirschberg and Manning(2015)}]{hirschberg2015advances}
\bibinfo{author}{Hirschberg, J.}, \bibinfo{author}{Manning, C.}, \bibinfo{year}{2015}.
\newblock \bibinfo{title}{Advances in natural language processing}.
\newblock \bibinfo{journal}{Science} \bibinfo{volume}{349}, \bibinfo{pages}{261--266}.
\newblock \DOIprefix\doi{10.1126/science.aaa8685}.
%Type = Misc
\bibitem[{Honnibal and Montani(2020)}]{honnibal2020}
\bibinfo{author}{Honnibal, M.}, \bibinfo{author}{Montani, I.}, \bibinfo{year}{2020}.
\newblock \bibinfo{title}{spacy: Industrial-strength natural language processing in python}.
\newblock \bibinfo{howpublished}{https://doi.org/10.5281/zenodo.1212303}.
%Type = Inproceedings
\bibitem[{Hutto and Gilbert(2014)}]{hutto2014vader}
\bibinfo{author}{Hutto, C.}, \bibinfo{author}{Gilbert, E.}, \bibinfo{year}{2014}.
\newblock \bibinfo{title}{Vader: A parsimonious rule-based model for sentiment analysis of social media text}, in: \bibinfo{booktitle}{Proceedings of the International AAAI Conference on Web and Social Media}, pp. \bibinfo{pages}{216--225}.
%Type = Article
\bibitem[{Ismagilova et~al.(2020)Ismagilova, Slade, Rana and Dwivedi}]{ismagilova2020effect}
\bibinfo{author}{Ismagilova, E.}, \bibinfo{author}{Slade, E.}, \bibinfo{author}{Rana, N.}, \bibinfo{author}{Dwivedi, Y.}, \bibinfo{year}{2020}.
\newblock \bibinfo{title}{The effect of characteristics of source credibility on consumer behaviour: A meta-analysis}.
\newblock \bibinfo{journal}{Journal of Retailing and Consumer Services} \bibinfo{volume}{53}, \bibinfo{pages}{101736}.
%Type = Inproceedings
\bibitem[{Jakesch et~al.(2019)Jakesch, French, Ma, Hancock and Naaman}]{jakesch2019ai-mediated}
\bibinfo{author}{Jakesch, M.}, \bibinfo{author}{French, M.}, \bibinfo{author}{Ma, X.}, \bibinfo{author}{Hancock, J.}, \bibinfo{author}{Naaman, M.}, \bibinfo{year}{2019}.
\newblock \bibinfo{title}{Ai-mediated communication: How the perception that profile text was written by ai affects trustworthiness}, in: \bibinfo{booktitle}{Proceedings of the 2019 CHI Conference on Human Factors in Computing Systems}, pp. \bibinfo{pages}{1--13}.
%Type = Article
\bibitem[{Jussupow et~al.(2020)Jussupow, Benbasat and Heinzl}]{jussupow2020why}
\bibinfo{author}{Jussupow, E.}, \bibinfo{author}{Benbasat, I.}, \bibinfo{author}{Heinzl, A.}, \bibinfo{year}{2020}.
\newblock \bibinfo{title}{Why are we averse towards algorithms? a comprehensive literature review on algorithm aversion}.
\newblock \bibinfo{journal}{ECIS 2020 Proceedings} .
%Type = Misc
\bibitem[{JustAnotherArchivist(2021)}]{snscrape2021}
\bibinfo{author}{JustAnotherArchivist}, \bibinfo{year}{2021}.
\newblock \bibinfo{title}{snscrape: A social networking service scraper in python}.
\newblock \bibinfo{howpublished}{Github}.
\newblock \URLprefix \url{https://github.com/JustAnotherArchivist/snscrape}.
%Type = Article
\bibitem[{Kaddour et~al.(2023)Kaddour, Harris, Mozes, Bradley, Raileanu and McHardy}]{kaddour2023challenges}
\bibinfo{author}{Kaddour, J.}, \bibinfo{author}{Harris, J.}, \bibinfo{author}{Mozes, M.}, \bibinfo{author}{Bradley, H.}, \bibinfo{author}{Raileanu, R.}, \bibinfo{author}{McHardy, R.}, \bibinfo{year}{2023}.
\newblock \bibinfo{title}{Challenges and applications of large language models}.
\newblock \bibinfo{journal}{arXiv} .
%Type = Article
\bibitem[{Karinshak et~al.(2023)Karinshak, Liu, Park and Hancock}]{karinshak2023working}
\bibinfo{author}{Karinshak, E.}, \bibinfo{author}{Liu, S.}, \bibinfo{author}{Park, J.}, \bibinfo{author}{Hancock, J.}, \bibinfo{year}{2023}.
\newblock \bibinfo{title}{Working with ai to persuade: Examining a large language model’s ability to generate pro-vaccination messages}.
\newblock \bibinfo{journal}{Proceedings of the ACM on Human-Computer Interaction (CSCW)} \bibinfo{volume}{7}, \bibinfo{pages}{1--29}.
%Type = Article
\bibitem[{Lagerkvist(2013)}]{lagerkvist2013consumer}
\bibinfo{author}{Lagerkvist, C.}, \bibinfo{year}{2013}.
\newblock \bibinfo{title}{Consumer preferences for food labelling attributes: Comparing direct ranking and best--worst scaling for measurement of attribute importance, preference intensity and attribute dominance}.
\newblock \bibinfo{journal}{Food Quality and Preference} \bibinfo{volume}{29}, \bibinfo{pages}{77--88}.
%Type = Incollection
\bibitem[{Lasswell(1948)}]{lasswell1948structure}
\bibinfo{author}{Lasswell, H.}, \bibinfo{year}{1948}.
\newblock \bibinfo{title}{The structure and function of communication in society}, in: \bibinfo{editor}{Bryson, L.} (Ed.), \bibinfo{booktitle}{The Communication of Ideas}. \bibinfo{publisher}{New York: Institute for Religious and Social Studies}, pp. \bibinfo{pages}{37--51}.
%Type = Article
\bibitem[{Lim and Schm{\"a}lzle(2023)}]{lim2023artificial}
\bibinfo{author}{Lim, S.}, \bibinfo{author}{Schm{\"a}lzle, R.}, \bibinfo{year}{2023}.
\newblock \bibinfo{title}{Artificial intelligence for health message generation: an empirical study using a large language model (llm) and prompt engineering}.
\newblock \bibinfo{journal}{Frontiers in Communication} \bibinfo{volume}{8}, \bibinfo{pages}{1129082}.
%Type = Article
\bibitem[{Liu and Yang(2020)}]{liu2020incorporating}
\bibinfo{author}{Liu, S.}, \bibinfo{author}{Yang, J.}, \bibinfo{year}{2020}.
\newblock \bibinfo{title}{Incorporating message framing into narrative persuasion to curb e‐cigarette use among college students}.
\newblock \bibinfo{journal}{Risk Analysis} \bibinfo{volume}{40}, \bibinfo{pages}{1677--1690}.
%Type = Inproceedings
\bibitem[{Liu et~al.(2022)Liu, Mittal, Yang and Bruckman}]{liu2022will}
\bibinfo{author}{Liu, Y.}, \bibinfo{author}{Mittal, A.}, \bibinfo{author}{Yang, D.}, \bibinfo{author}{Bruckman, A.}, \bibinfo{year}{2022}.
\newblock \bibinfo{title}{Will ai console me when i lose my pet? understanding perceptions of ai-mediated email writing}, in: \bibinfo{booktitle}{Proceedings of the 2022 CHI conference on human factors in computing systems}, pp. \bibinfo{pages}{1--13}.
%Type = Article
\bibitem[{Longoni et~al.(2019)Longoni, Bonezzi and Morewedge}]{longoni2019resistance}
\bibinfo{author}{Longoni, C.}, \bibinfo{author}{Bonezzi, A.}, \bibinfo{author}{Morewedge, C.}, \bibinfo{year}{2019}.
\newblock \bibinfo{title}{Resistance to medical artificial intelligence}.
\newblock \bibinfo{journal}{Journal of Consumer Research} \bibinfo{volume}{46}, \bibinfo{pages}{629--650}.
%Type = Book
\bibitem[{Luger(2005)}]{luger2005artificial}
\bibinfo{author}{Luger, G.}, \bibinfo{year}{2005}.
\newblock \bibinfo{title}{Artificial Intelligence: Structures and strategies for complex problem solving}.
\newblock \bibinfo{publisher}{London: Pearson Education}.
%Type = Article
\bibitem[{Lyu et~al.(2021)Lyu, Luli and Ling}]{lyu2021vaping}
\bibinfo{author}{Lyu, J.}, \bibinfo{author}{Luli, G.}, \bibinfo{author}{Ling, P.}, \bibinfo{year}{2021}.
\newblock \bibinfo{title}{Vaping discussion in the covid-19 pandemic: An observational study using twitter data}.
\newblock \bibinfo{journal}{PloS One} \bibinfo{volume}{16}, \bibinfo{pages}{e0260290}.
%Type = Article
\bibitem[{Ma and Atkin(2017)}]{ma2017user}
\bibinfo{author}{Ma, T.}, \bibinfo{author}{Atkin, D.}, \bibinfo{year}{2017}.
\newblock \bibinfo{title}{User generated content and credibility evaluation of online health information: A meta analytic study}.
\newblock \bibinfo{journal}{Telematics and Informatics} \bibinfo{volume}{34}, \bibinfo{pages}{472--486}.
%Type = Article
\bibitem[{Marcus(2018)}]{marcus2018deep}
\bibinfo{author}{Marcus, G.}, \bibinfo{year}{2018}.
\newblock \bibinfo{title}{Deep learning: A critical appraisal}.
\newblock \bibinfo{journal}{arXiv} .
%Type = Article
\bibitem[{Marcus(2020)}]{marcus2020next}
\bibinfo{author}{Marcus, G.}, \bibinfo{year}{2020}.
\newblock \bibinfo{title}{The next decade in ai: four steps towards robust artificial intelligence}.
\newblock \bibinfo{journal}{arXiv} \URLprefix \url{https://doi.org/10.48550/arXiv.2002.06177}.
%Type = Article
\bibitem[{Miles et~al.(2021)Miles, West and Nadarzynski}]{miles2021health}
\bibinfo{author}{Miles, O.}, \bibinfo{author}{West, R.}, \bibinfo{author}{Nadarzynski, T.}, \bibinfo{year}{2021}.
\newblock \bibinfo{title}{Health chatbots acceptability moderated by perceived stigma and severity: A cross-sectional survey}.
\newblock \bibinfo{journal}{Digital Health} \bibinfo{volume}{7}, \bibinfo{pages}{20552076211063012}.
%Type = Book
\bibitem[{Mitchell(2019)}]{mitchell2019artificial}
\bibinfo{author}{Mitchell, M.}, \bibinfo{year}{2019}.
\newblock \bibinfo{title}{Artificial Intelligence: A guide for thinking humans}.
\newblock \bibinfo{publisher}{Penguin UK}, \bibinfo{address}{London}.
%Type = Book
\bibitem[{Murphy(1981)}]{murphy1981rhetoric}
\bibinfo{author}{Murphy, J.J.}, \bibinfo{year}{1981}.
\newblock \bibinfo{title}{Rhetoric in the Middle Ages: A history of rhetorical theory from Saint Augustine to the Renaissance}.
\newblock \bibinfo{publisher}{University of California Press}, \bibinfo{address}{Berkeley, CA}.
%Type = Book
\bibitem[{Nahon and Hemsley(2013)}]{nahon2013going}
\bibinfo{author}{Nahon, K.}, \bibinfo{author}{Hemsley, J.}, \bibinfo{year}{2013}.
\newblock \bibinfo{title}{Going viral}.
\newblock \bibinfo{publisher}{Polity}.
%Type = Article
\bibitem[{Noar et~al.(2020)Noar, Rohde, Prentice-Dunn, Kresovich, Hall and Brewer}]{noar2020evaluating}
\bibinfo{author}{Noar, S.M.}, \bibinfo{author}{Rohde, J.A.}, \bibinfo{author}{Prentice-Dunn, H.}, \bibinfo{author}{Kresovich, A.}, \bibinfo{author}{Hall, M.G.}, \bibinfo{author}{Brewer, N.T.}, \bibinfo{year}{2020}.
\newblock \bibinfo{title}{Evaluating the actual and perceived effectiveness of e-cigarette prevention advertisements among adolescents}.
\newblock \bibinfo{journal}{Addictive Behaviors} \bibinfo{volume}{109}, \bibinfo{pages}{106473}.
%Type = Book
\bibitem[{O'Keefe(2015)}]{okeefe2015persuasion}
\bibinfo{author}{O'Keefe, D.J.}, \bibinfo{year}{2015}.
\newblock \bibinfo{title}{Persuasion: Theory and research}.
\newblock \bibinfo{publisher}{Sage Publications}.
%Type = Article
\bibitem[{Ozawa et~al.(2017)Ozawa, Wonodi, Babalola, Ismail and Bridges}]{ozawa2017using}
\bibinfo{author}{Ozawa, S.}, \bibinfo{author}{Wonodi, C.}, \bibinfo{author}{Babalola, O.}, \bibinfo{author}{Ismail, T.}, \bibinfo{author}{Bridges, J.}, \bibinfo{year}{2017}.
\newblock \bibinfo{title}{Using best-worst scaling to rank factors affecting vaccination demand in northern nigeria}.
\newblock \bibinfo{journal}{Vaccine} \bibinfo{volume}{35}, \bibinfo{pages}{6429--6437}.
%Type = Article
\bibitem[{Palan and Schitter(2018)}]{palan2018prolific}
\bibinfo{author}{Palan, S.}, \bibinfo{author}{Schitter, C.}, \bibinfo{year}{2018}.
\newblock \bibinfo{title}{Prolific. ac—a subject pool for online experiments}.
\newblock \bibinfo{journal}{Journal of Behavioral and Experimental Finance} \bibinfo{volume}{17}, \bibinfo{pages}{22--27}.
%Type = Inproceedings
\bibitem[{Petty and Cacioppo(1986)}]{petty1986elaboration}
\bibinfo{author}{Petty, R.E.}, \bibinfo{author}{Cacioppo, J.T.}, \bibinfo{year}{1986}.
\newblock \bibinfo{title}{The elaboration likelihood model of persuasion}, in: \bibinfo{booktitle}{Springer New York}, pp. \bibinfo{pages}{1--24}.
%Type = Article
\bibitem[{Pornpitakpan(2004)}]{pornpitakpan2004persuasiveness}
\bibinfo{author}{Pornpitakpan, C.}, \bibinfo{year}{2004}.
\newblock \bibinfo{title}{The persuasiveness of source credibility: A critical review of five decades' evidence}.
\newblock \bibinfo{journal}{Journal of Applied Social Psychology} \bibinfo{volume}{34}, \bibinfo{pages}{243--281}.
%Type = Article
\bibitem[{Prasetyo et~al.(2021)Prasetyo, Dewi, Balatbat, Antonio, Chuenyindee, Perwira~Redi, Young, Diaz and Kurata}]{prasetyo2021evaluation}
\bibinfo{author}{Prasetyo, Y.T.}, \bibinfo{author}{Dewi, R.S.}, \bibinfo{author}{Balatbat, N.M.}, \bibinfo{author}{Antonio, M.L.B.}, \bibinfo{author}{Chuenyindee, T.}, \bibinfo{author}{Perwira~Redi, A.A.N.}, \bibinfo{author}{Young, M.N.}, \bibinfo{author}{Diaz, J.F.T.}, \bibinfo{author}{Kurata, Y.B.}, \bibinfo{year}{2021}.
\newblock \bibinfo{title}{The evaluation of preference and perceived quality of health communication icons associated with covid-19 prevention measures}.
\newblock \bibinfo{journal}{Healthcare} \bibinfo{volume}{9}, \bibinfo{pages}{1115}.
%Type = Inproceedings
\bibitem[{Ragot et~al.(2020)Ragot, Martin and Cojean}]{ragot2020ai}
\bibinfo{author}{Ragot, M.}, \bibinfo{author}{Martin, N.}, \bibinfo{author}{Cojean, S.}, \bibinfo{year}{2020}.
\newblock \bibinfo{title}{Ai-generated vs. human artworks. a perception bias towards artificial intelligence?}, in: \bibinfo{booktitle}{Extended abstracts of the 2020 CHI conference on human factors in computing systems}, pp. \bibinfo{pages}{1--10}.
%Type = Article
\bibitem[{Ratneshwar and Chaiken(1991)}]{ratneshwar1991comprehension}
\bibinfo{author}{Ratneshwar, S.}, \bibinfo{author}{Chaiken, S.}, \bibinfo{year}{1991}.
\newblock \bibinfo{title}{Comprehension's role in persuasion: The case of its moderating effect on the persuasive impact of source cues}.
\newblock \bibinfo{journal}{Journal of Consumer Research} \bibinfo{volume}{18}, \bibinfo{pages}{52--62}.
%Type = Article
\bibitem[{Reimers and Gurevych(2019)}]{reimers2019sentence}
\bibinfo{author}{Reimers, N.}, \bibinfo{author}{Gurevych, I.}, \bibinfo{year}{2019}.
\newblock \bibinfo{title}{Sentence-bert: Sentence embeddings using siamese bert-networks}.
\newblock \bibinfo{journal}{arXiv} \URLprefix \url{http://arxiv.org/abs/1908.10084}.
%Type = Article
\bibitem[{Rohde et~al.(2021)Rohde, Noar, Prentice-Dunn, Kresovich and Hall}]{rohde2021comparison}
\bibinfo{author}{Rohde, J.A.}, \bibinfo{author}{Noar, S.M.}, \bibinfo{author}{Prentice-Dunn, H.}, \bibinfo{author}{Kresovich, A.}, \bibinfo{author}{Hall, M.G.}, \bibinfo{year}{2021}.
\newblock \bibinfo{title}{Comparison of message and effects perceptions for the real cost e-cigarette prevention ads}.
\newblock \bibinfo{journal}{Health Communication} \bibinfo{volume}{36}, \bibinfo{pages}{1222--1230}.
%Type = Book
\bibitem[{Russell and Norvig(2021)}]{russell2021artificial}
\bibinfo{author}{Russell, S.}, \bibinfo{author}{Norvig, P.}, \bibinfo{year}{2021}.
\newblock \bibinfo{title}{Artificial intelligence: A modern approach 4th Edition}.
\newblock \bibinfo{publisher}{Prentice Hall}, \bibinfo{address}{Hoboken}.
%Type = Article
\bibitem[{Scao et~al.(2022)Scao, Fan, Akiki, Pavlick, Ili{\'c}, Hesslow and Manica}]{scao2022bloom}
\bibinfo{author}{Scao, T.}, \bibinfo{author}{Fan, A.}, \bibinfo{author}{Akiki, C.}, \bibinfo{author}{Pavlick, E.}, \bibinfo{author}{Ili{\'c}, S.}, \bibinfo{author}{Hesslow, D.}, \bibinfo{author}{Manica, M.}, \bibinfo{year}{2022}.
\newblock \bibinfo{title}{Bloom: A 176b-parameter open-access multilingual language model}.
\newblock \bibinfo{journal}{arXiv preprint arXiv:2211.05100} .
%Type = Article
\bibitem[{Schepman and Rodway(2023)}]{schepman2023general}
\bibinfo{author}{Schepman, A.}, \bibinfo{author}{Rodway, P.}, \bibinfo{year}{2023}.
\newblock \bibinfo{title}{The general attitudes towards artificial intelligence scale (gaais): Confirmatory validation and associations with personality, corporate distrust, and general trust}.
\newblock \bibinfo{journal}{International Journal of Human-Computer Interaction} \bibinfo{volume}{39}, \bibinfo{pages}{2724--2741}.
%Type = Article
\bibitem[{Schm{\"a}lzle and Wilcox(2022)}]{schmalzle2022harnessing}
\bibinfo{author}{Schm{\"a}lzle, R.}, \bibinfo{author}{Wilcox, S.}, \bibinfo{year}{2022}.
\newblock \bibinfo{title}{Harnessing artificial intelligence for health message generation: The folic acid message engine}.
\newblock \bibinfo{journal}{Journal of Medical Internet Research} \bibinfo{volume}{24}, \bibinfo{pages}{e28858}.
%Type = Article
\bibitem[{Shank et~al.(2023)Shank, Stefanik, Stuhlsatz, Kacirek and Belfi}]{shank2023ai}
\bibinfo{author}{Shank, D.B.}, \bibinfo{author}{Stefanik, C.}, \bibinfo{author}{Stuhlsatz, C.}, \bibinfo{author}{Kacirek, K.}, \bibinfo{author}{Belfi, A.M.}, \bibinfo{year}{2023}.
\newblock \bibinfo{title}{Ai composer bias: Listeners like music less when they think it was composed by an ai}.
\newblock \bibinfo{journal}{Journal of Experimental Psychology: Applied} \bibinfo{volume}{29}, \bibinfo{pages}{676}.
%Type = Article
\bibitem[{Shannon(1948)}]{shannon1948mathematical}
\bibinfo{author}{Shannon, C.}, \bibinfo{year}{1948}.
\newblock \bibinfo{title}{A mathematical theory of communication}.
\newblock \bibinfo{journal}{Bell Systems Technical Journal} \bibinfo{volume}{27}, \bibinfo{pages}{379--423}.
%Type = Article
\bibitem[{Slovic(1987)}]{slovic1987psychometric}
\bibinfo{author}{Slovic, P.}, \bibinfo{year}{1987}.
\newblock \bibinfo{title}{The psychometric paradigm}.
\newblock \bibinfo{journal}{Science} \bibinfo{volume}{236}, \bibinfo{pages}{280--285}.
%Type = Book
\bibitem[{Tunstall et~al.(2022)Tunstall, von Werra and Wolf}]{tunstall2022natural}
\bibinfo{author}{Tunstall, L.}, \bibinfo{author}{von Werra, L.}, \bibinfo{author}{Wolf, T.}, \bibinfo{year}{2022}.
\newblock \bibinfo{title}{Natural language processing with Transformers}.
\newblock \bibinfo{publisher}{O’Reilly Media, Inc.}
%Type = Article
\bibitem[{Villanti et~al.(2021)Villanti, LePine, West, Cruz, Stevens, Tetreault and Mays}]{villanti2021identifying}
\bibinfo{author}{Villanti, A.C.}, \bibinfo{author}{LePine, S.E.}, \bibinfo{author}{West, J.C.}, \bibinfo{author}{Cruz, T.B.}, \bibinfo{author}{Stevens, E.M.}, \bibinfo{author}{Tetreault, H.J.}, \bibinfo{author}{Mays, D.}, \bibinfo{year}{2021}.
\newblock \bibinfo{title}{Identifying message content to reduce vaping: Results from online message testing trials in young adult tobacco users}.
\newblock \bibinfo{journal}{Addictive Behaviors} \bibinfo{volume}{115}, \bibinfo{pages}{106778}.
%Type = Article
\bibitem[{Wang et~al.(2021)Wang, Gentzke, Neff, Glidden, Jamal, Park-Lee and Hacker}]{wang2021characteristics}
\bibinfo{author}{Wang, T.W.}, \bibinfo{author}{Gentzke, A.S.}, \bibinfo{author}{Neff, L.J.}, \bibinfo{author}{Glidden, E.V.}, \bibinfo{author}{Jamal, A.}, \bibinfo{author}{Park-Lee, E.}, \bibinfo{author}{Hacker, K.A.}, \bibinfo{year}{2021}.
\newblock \bibinfo{title}{Characteristics of e-cigarette use behaviors among us youth, 2020}.
\newblock \bibinfo{journal}{JAMA Network Open} \bibinfo{volume}{4}, \bibinfo{pages}{e2111336--e2111336}.
%Type = Article
\bibitem[{Wang et~al.(2023)Wang, Xu, Wu, Kim, Fetterman, Hong and McLaughlin}]{wang2023moralization}
\bibinfo{author}{Wang, Y.}, \bibinfo{author}{Xu, Y.A.}, \bibinfo{author}{Wu, J.}, \bibinfo{author}{Kim, H.M.}, \bibinfo{author}{Fetterman, J.L.}, \bibinfo{author}{Hong, T.}, \bibinfo{author}{McLaughlin, M.L.}, \bibinfo{year}{2023}.
\newblock \bibinfo{title}{Moralization of e-cigarette use and regulation: A mixed-method computational analysis of opinion polarization}.
\newblock \bibinfo{journal}{Health Communication} \bibinfo{volume}{38}, \bibinfo{pages}{1666--1676}.
%Type = Article
\bibitem[{Wei et~al.(2022)Wei, Wang, Schuurmans, Bosma, Xia, Chi, V.~Le and Zhou}]{wei2022chain}
\bibinfo{author}{Wei, J.}, \bibinfo{author}{Wang, X.}, \bibinfo{author}{Schuurmans, D.}, \bibinfo{author}{Bosma, M.}, \bibinfo{author}{Xia, F.}, \bibinfo{author}{Chi, E.H.}, \bibinfo{author}{V.~Le, Q.}, \bibinfo{author}{Zhou, D.}, \bibinfo{year}{2022}.
\newblock \bibinfo{title}{Chain-of-thought prompting elicits reasoning in large language models}.
\newblock \bibinfo{journal}{Advances in Neural Information Processing Systems} \bibinfo{volume}{35}, \bibinfo{pages}{24824--24837}.
%Type = Article
\bibitem[{Wilson and Sherrell(1993)}]{wilson1993source}
\bibinfo{author}{Wilson, E.J.}, \bibinfo{author}{Sherrell, D.L.}, \bibinfo{year}{1993}.
\newblock \bibinfo{title}{Source effects in communication and persuasion research: A meta-analysis of effect size}.
\newblock \bibinfo{journal}{Journal of the Academy of Marketing Science} \bibinfo{volume}{21}, \bibinfo{pages}{101--112}.
%Type = Article
\bibitem[{Zhang et~al.(2023)Zhang, Li, Cui, Cai, Liu, Fu, Huang, Zhao, Zhang, Chen, Wang, Luu, Bi, Shi and Shi}]{zhang2023siren}
\bibinfo{author}{Zhang, Y.}, \bibinfo{author}{Li, Y.}, \bibinfo{author}{Cui, L.}, \bibinfo{author}{Cai, D.}, \bibinfo{author}{Liu, L.}, \bibinfo{author}{Fu, T.}, \bibinfo{author}{Huang, X.}, \bibinfo{author}{Zhao, E.}, \bibinfo{author}{Zhang, Y.}, \bibinfo{author}{Chen, Y.}, \bibinfo{author}{Wang, L.}, \bibinfo{author}{Luu, A.T.}, \bibinfo{author}{Bi, W.}, \bibinfo{author}{Shi, F.}, \bibinfo{author}{Shi, S.}, \bibinfo{year}{2023}.
\newblock \bibinfo{title}{Siren's song in the ai ocean: A survey on hallucination in large language models}.
\newblock \bibinfo{journal}{arXiv} \URLprefix \url{https://doi.org/10.48550/arXiv.2309.01219}.
%Type = Book
\bibitem[{Zhang and Yuan(2018)}]{zhang2018practical}
\bibinfo{author}{Zhang, Z.}, \bibinfo{author}{Yuan, K.H.}, \bibinfo{year}{2018}.
\newblock \bibinfo{title}{Practical statistical power analysis using Webpower and R}.
\newblock \bibinfo{publisher}{ISDSA Press}.
%Type = Article
\bibitem[{Zhou et~al.(2023)Zhou, Silvasstar, Clark, Salyers, Chavez and Bull}]{zhou2023ai}
\bibinfo{author}{Zhou, S.}, \bibinfo{author}{Silvasstar, J.}, \bibinfo{author}{Clark, C.}, \bibinfo{author}{Salyers, A.J.}, \bibinfo{author}{Chavez, C.}, \bibinfo{author}{Bull, S.S.}, \bibinfo{year}{2023}.
\newblock \bibinfo{title}{An artificially intelligent, natural language processing chatbot designed to promote covid-19 vaccination: A proof-of-concept pilot study}.
\newblock \bibinfo{journal}{Digital Health} \bibinfo{volume}{9}, \bibinfo{pages}{20552076231155679}.

\end{thebibliography}

\end{document}